\documentclass[letterpaper,11pt,conference,onecolumn]{ieeeconf}

\IEEEoverridecommandlockouts
\usepackage{graphicx}
\usepackage{fontawesome5}
\graphicspath{{./}}
\usepackage{amsmath,amssymb}
\makeatletter
\let\NAT@parse\undefined
\makeatother
\usepackage[authoryear,round]{natbib}
\setcitestyle{aysep={,},yysep={,},semicolon}

\usepackage{booktabs}
\usepackage{multirow}
\usepackage{microtype}
\usepackage{xcolor}
\usepackage{colortbl}
\usepackage{tabularx}
\usepackage{xspace}
\usepackage{placeins}
\usepackage{wrapfig}
\usepackage{url}
\usepackage[hidelinks]{hyperref}
\usepackage{tcolorbox}
\usepackage[scaled=0.95]{helvet}
\definecolor{logivtitlebg}{HTML}{F0EAF7}
\definecolor{logivlink}{HTML}{7952B3}
\hypersetup{colorlinks=true,citecolor=logivlink,linkcolor=logivlink,urlcolor=logivlink}
\AtBeginDocument{%
  \NewCommandCopy{\logivOriginalCite}{\citep}
  \RenewDocumentCommand{\cite}{o m}{%
    \IfNoValueTF{#1}
      {{\color{logivlink}\logivOriginalCite{#2}}}
      {{\color{logivlink}\logivOriginalCite[#1]{#2}}}}}

\newcommand{\method}{LOGIV\xspace}

\definecolor{algcommentblue}{RGB}{0,125,155}
\newcounter{algorithm}
\newcounter{algline}
\newenvironment{algorithm}[1][t]{%
  \begin{figure}[#1]\refstepcounter{algorithm}\footnotesize\hrule\vspace{3pt}
}{\end{figure}}
\newcommand{\algcaption}[1]{%
  \noindent\textbf{Algorithm~\thealgorithm\quad #1}\par\vspace{3pt}\hrule\vspace{4pt}}
\newenvironment{compactalgorithmic}{%
  \begin{list}{\arabic{algline}}{%
    \usecounter{algline}
    \setlength{\leftmargin}{1.55em}
    \setlength{\labelwidth}{1.05em}
    \setlength{\labelsep}{0.45em}
    \setlength{\itemsep}{0.8pt}
    \setlength{\parsep}{0pt}
    \setlength{\topsep}{3pt}
    \setlength{\partopsep}{0pt}}
}{\end{list}\vspace{3pt}\hrule\vspace{3pt}}
\newcommand{\algcomment}[1]{{\color{algcommentblue}// #1}}

\title{LOGIV: Graph-Verified Execution with Causal Local Repair for Long-Horizon Robot Tasks}
\author{Enqiao Lu \and Yuxiang Xie \and Xingrui Yu \and Yuyang Wang \and Yipeng Gu \and Zhirui Fang \and Xinrun Xu \and Zhenglin Wan \and Pengfei Zhou \and Wangbo Zhao \and Yang You \and Yuxiang Gao \and Ivor Tsang}

\begin{document}
\raggedbottom
\begin{tcolorbox}[colback=logivtitlebg,colframe=logivtitlebg,
  boxrule=0pt,arc=10pt,left=14pt,right=14pt,top=14pt,bottom=12pt,
  before skip=0pt,after skip=14pt]
{\sffamily\bfseries\fontsize{20}{23}\selectfont
LOGIV: Graph-Verified Execution with Causal Local Repair for Long-Horizon Robot Tasks\par}
\vspace{8pt}
{\sffamily\bfseries\fontsize{10}{13}\selectfont\raggedright
Enqiao Lu\textsuperscript{1,2,3,*},
Yuxiang Xie\textsuperscript{1,*},
Xingrui Yu\textsuperscript{2,\textdagger},
Yuyang Wang\textsuperscript{4},
Yipeng Gu\textsuperscript{1},
Zhirui Fang\textsuperscript{3,7},
Xinrun Xu\textsuperscript{3},
Zhenglin Wan\textsuperscript{5},
Pengfei Zhou\textsuperscript{5},
Wangbo Zhao\textsuperscript{6},
Yang You\textsuperscript{5},
Yuxiang Gao,
Ivor Tsang\textsuperscript{2,7}\par}
\vspace{5pt}
{\fontsize{10}{12}\selectfont
\textsuperscript{1}CUHKSZ, \textsuperscript{2}A*STAR,
\textsuperscript{3}THU, \textsuperscript{4}Caltech,
\textsuperscript{5}NUS, \textsuperscript{6}HKUST, \textsuperscript{7}NTU\par
\textsuperscript{*}Equal contribution, \textsuperscript{\textdagger}Corresponding author\par}
\vspace{12pt}
\begingroup\fontsize{10.5}{13}\selectfont
\thispagestyle{empty}
\pagestyle{empty}

Agentic robots must turn a high-level goal into actions while keeping the physical conditions needed for future steps valid. A completed action does not ensure that its effects will persist: a later disturbance can invalidate support that the remaining plan still assumes. We present LOGIV, a graph engineering framework for agentic robot execution that uses an explicit task graph to connect symbolic planning, observation-based verification, and embodied action. LOGIV compiles a validated plan into condition-support and resource-ordering edges, checks their physical validity during execution, and uses dependency tracing to determine what must be restored after a failure. Repair is constrained by retained unfinished actions, and the combined continuation is symbolically validated before execution resumes. The framework operates above existing Vision-Language-Action (VLA) and World Action Model (WAM) policies without adding trainable components. Across 20 dual-arm tasks from RoboTwin~V2 and BiCoord with three execution backbones, LOGIV improves mean task success without injected disturbances by 10.67 and 15.33 percentage points, respectively. Under predefined perturbations, it achieves higher task completion and condition restoration than three adapted recovery baselines for every backbone on both benchmarks. On three real-robot tasks, successful trials increase from 12/30 to 17/30. These results support explicit, continuously checked task dependencies as a practical interface for reliable agentic robot execution.\par\endgroup
\vspace{7pt}
{\normalfont\rmfamily\fontsize{10}{13}\selectfont
\urlstyle{same}\raggedright
\faEnvelope\enspace Email: \href{mailto:124090411@link.cuhk.edu.cn}{\nolinkurl{124090411@link.cuhk.edu.cn}},\enspace\href{mailto:yu_xingrui@a-star.edu.sg}{\nolinkurl{yu_xingrui@a-star.edu.sg}}\par
\faGithub\enspace GitHub: \url{https://github.com/Enqiao-Lu/LOGIV}\par
\textcolor{orange}{\faFire}\enspace Project Page: \url{https://enqiao-lu.github.io/logiv-project/}\par
}
\vspace{16pt}
\noindent\hfill
\raisebox{-0.5\height}{\includegraphics[height=22pt,trim=0bp 0bp 571bp 0bp,clip]{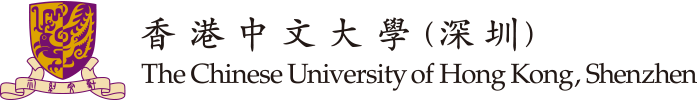}}\hspace{10pt}
\raisebox{-0.5\height}{\includegraphics[height=22pt,trim=3.84bp 6.24bp 72.48bp 5.52bp,clip]{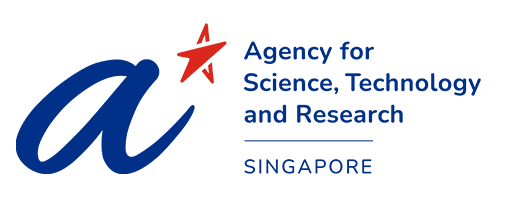}}\hspace{10pt}
\raisebox{-0.5\height}{\includegraphics[height=22pt]{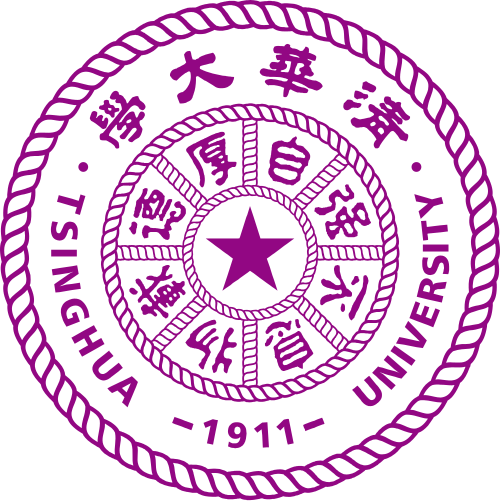}}\hspace{10pt}
\raisebox{-0.5\height}{\includegraphics[height=22pt]{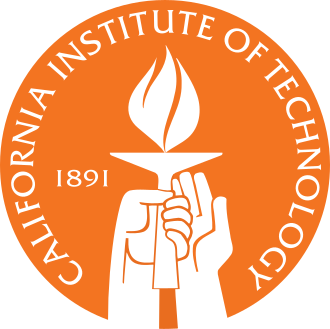}}\hspace{10pt}
\raisebox{-0.5\height}{\includegraphics[height=22pt,trim=0bp 0bp 46.08bp 0bp,clip]{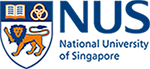}}\hspace{10pt}
\raisebox{-0.5\height}{\includegraphics[height=22pt,trim=0bp 0bp 201bp 0bp,clip]{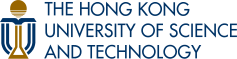}}\hspace{10pt}
\raisebox{-0.5\height}{\includegraphics[height=22pt,trim=0bp 0bp 520bp 0bp,clip]{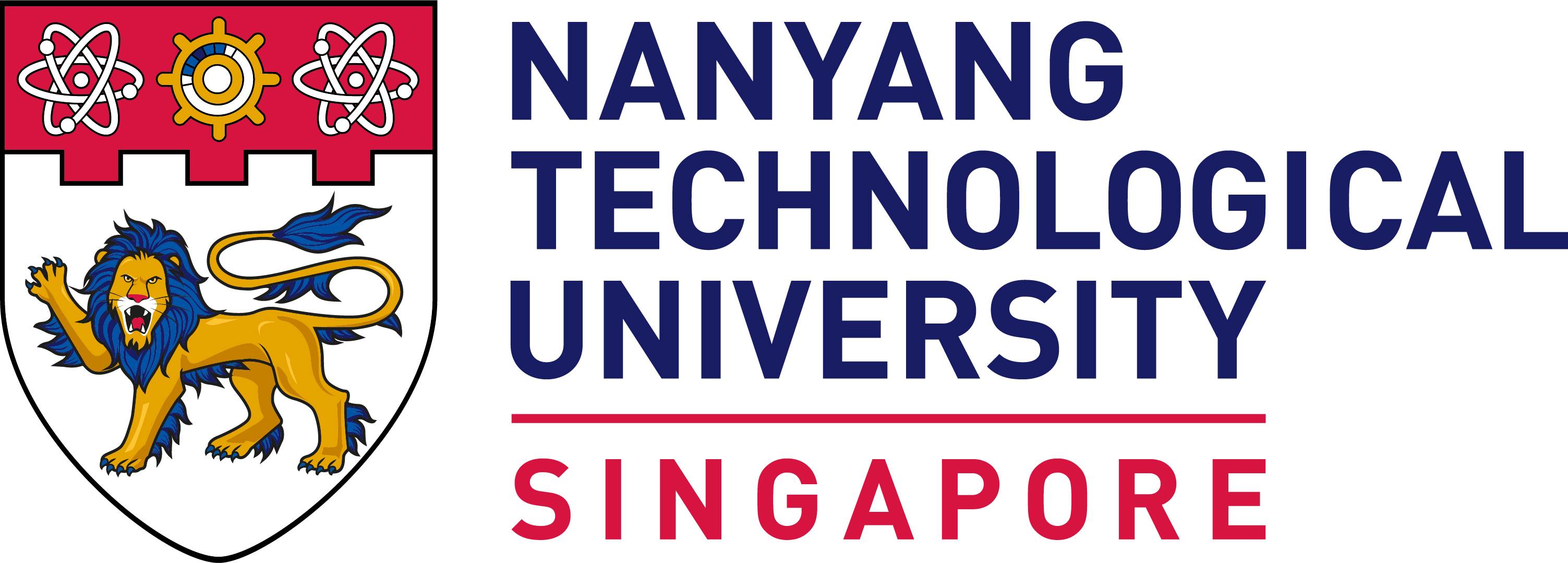}}
\end{tcolorbox}

\section{Introduction}
\begin{wrapfigure}{R}{0.48\textwidth}
  \centering
  \includegraphics[width=\linewidth]{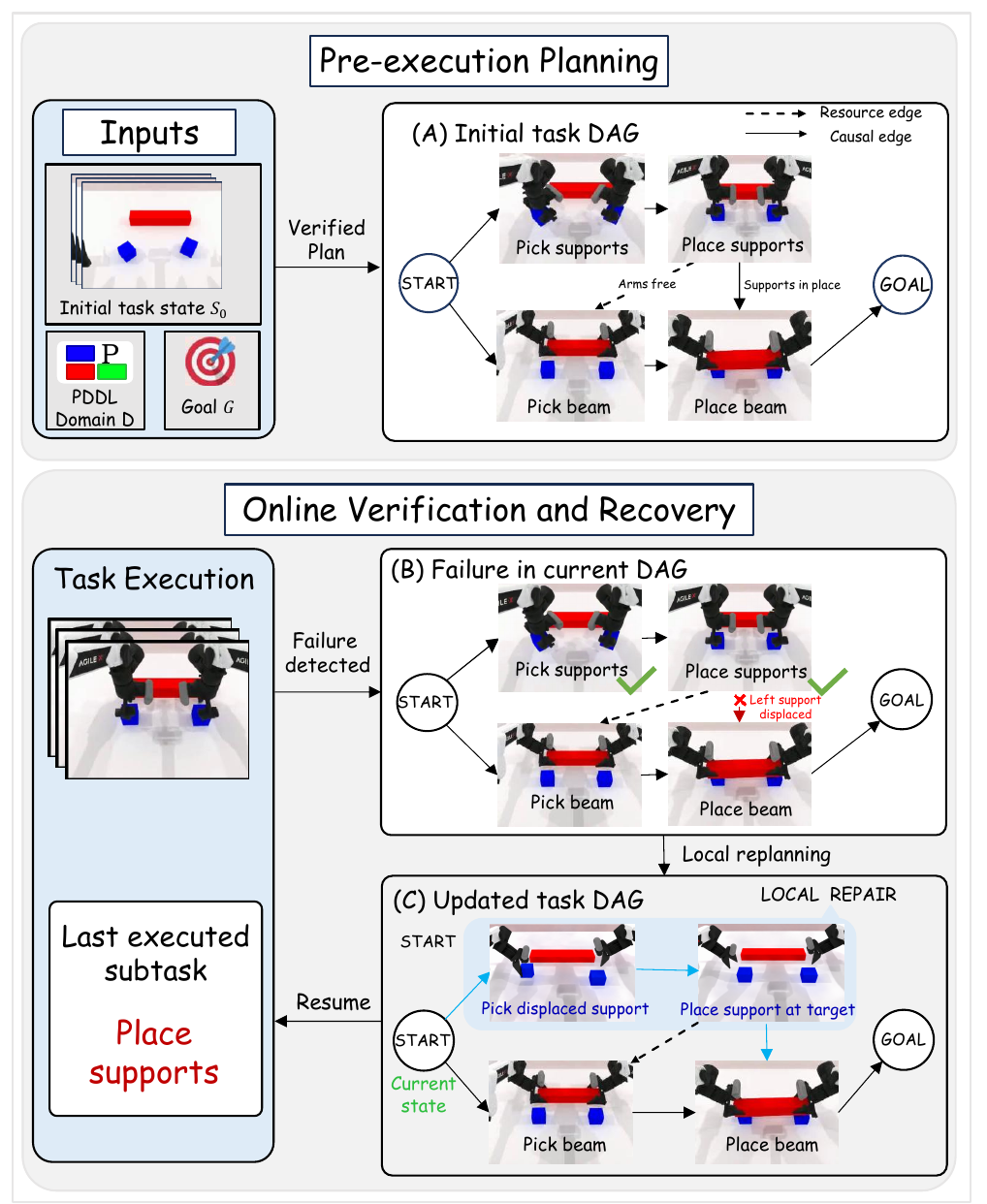}
  \caption{\method execution on the BiCoord \texttt{build\_bridge} task.
  (A) A validated reference plan is compiled into the initial task DAG before execution.
  (B) A previously placed support is displaced, invalidating a condition required by the remaining plan.
  (C) \method repairs only the affected support operations, retains the unaffected beam operations, and resumes execution on the updated task DAG.}
  \label{fig:teaser}
\end{wrapfigure}

Agentic robots combine goal-directed planning, perception, and physical action in a feedback loop~\cite{qin2026agentic}. Vision-Language-Action (VLA) models~\cite{team2024octo,kim2024openvla,intelligence2025pi_} and World Action Models (WAMs)~\cite{yuan2026fast,cai2026aha} provide increasingly capable execution policies, while an agent-level controller must decide how their actions contribute to a longer task. A distinctive difficulty in physical execution is that an intermediate result can cease to hold after the action that produced it has succeeded. A support placed during bridge assembly may be displaced before the beam is installed (Fig.~\ref{fig:teaser}). The placement remains a completed action, yet the support condition needed by the remaining plan is no longer valid. Reliable long-horizon behavior therefore requires tracking both action completion and the continuing validity of its consequences.

Related progress spans structured agent capabilities~\cite{yang2026federated}, multi-agent interaction topology~\cite{chen-etal-2026-diversity}, and feedback-based visual refinement~\cite{zhang2026closed}. LLM- and VLM-based systems select skills, interpret feedback, and organize multistep behavior~\cite{ouyang2025long,huang2024rekep,feng2025reflective,chen2025robohorizon,ji2026hime,peng2026cortex,zhang2026harness,huang2026roboharness}. Symbolic and program-based methods make action constraints explicit~\cite{yang2025guiding,kwon2025fast,chen2025code,chu2025llm+}, building on established work in plan monitoring and repair~\cite{pollack1999towards,van2005plan}. Graph-based agent frameworks expose subtask dependencies for reuse and repair~\cite{zhang2026atomic}; in robotics, AgentChord augments task graphs with anticipatory recovery branches before execution~\cite{xu2026agentchord}. These advances raise a question: when observations invalidate an earlier condition, which support should the robot restore while preserving a usable unfinished plan?

We approach this question through \emph{graph engineering}: constructing, checking, and revising an explicit dependency graph as the interface between planning and robot execution. We present \method(\underline{L}ong-Horizon \underline{O}nline \underline{G}raph-based \underline{I}nference and \underline{V}erification for Plan Repair). Its task graph records which conditions each action supplies to later actions, together with resource-ordering constraints. The key design is to separate a node's completion status from the current validity of the support it provides. When that support fails, unfinished consumers can remain in the plan while repair re-establishes their prerequisites. The retained continuation then constrains the repair goal. This gives graph structure an operational role in deciding what to check, what to preserve, and what to change.

LOGIV implements this interface as a closed-loop execution and repair layer above an existing VLA or WAM policy. Before execution, it validates a high-level plan and compiles its dependencies into a task graph. During execution, visual observations and robot feedback check the selected action and the support required by pending actions. After a confirmed failure, dependency tracing identifies affected actions; repair synthesizes a continuation constrained by retained operators and expands the replacement scope when validation fails. The combined plan is checked under the symbolic model before the graph is rebuilt and execution resumes. In the bridge example, the robot restores the displaced support while retaining the remaining beam operations. This design adds no trainable components and focuses on online repair from the observed state, with the validity of the repaired continuation checked explicitly. We evaluate LOGIV on 20 simulated dual-arm tasks from RoboTwin~V2 and BiCoord using $\pi_{0.5}$, AHA-WAM, and Light-WAM. Without injected perturbations, mean success across the three backbones increases from 71.67\% to 82.33\% on RoboTwin~V2 and from 42.33\% to 57.67\% on BiCoord. Under predefined perturbations, LOGIV achieves higher task completion and condition restoration than three adapted recovery baselines for every backbone on both benchmarks. On three real-robot tasks, successful trials increase from 12/30 to 17/30. Ablations support the role of graph structure: removing dependency checks reduces perturbed-task success from 60\% to 48\%, and replacing local repair with replanning all remaining actions increases repeated valid actions from 0.2 to 1.6. These results support graph engineering for the evaluated tasks. Our contributions are:
\begin{itemize}
  \setlength{\itemsep}{0pt}
  \setlength{\topsep}{2pt}
  \item A graph engineering formulation of agentic robot execution that separates completed actions from the continuing validity of their condition support during physical task execution.\looseness=-1
  \item A closed-loop implementation combining observation-based graph checks, dependency-guided repair constrained by retained actions, and symbolic validation of the full continuation.
  \item Evaluation across two simulation benchmarks, three execution backbones, and real robots, with ablations examining how graph verification and repair design affect task success and redundant execution.
\end{itemize}

\section{Related Work}
\subsection{Long-Horizon Manipulation with VLA and WAM}

Recent approaches improve long-horizon manipulation through high-level execution organization or changes to the policy backbone. In the first group, a high-level module selects and sequences subtasks while a low-level policy executes them. HiMe uses hierarchical memory to preserve task context, Cortex aligns planning and execution interfaces, and Harness VLA and RoboHarness use execution memory to organize robot policies~\cite{ji2026hime,peng2026cortex,zhang2026harness,huang2026roboharness}. EMERGE-Policy coordinates specialized agents for perception, monitoring, verification, and memory through a graph-structured framework with a common skill interface~\cite{fang2026emerge}. These methods motivate the use of an explicit execution layer around an existing policy. Other approaches improve the representation or use of temporal information within the model. RoboHorizon combines multiview world modeling with mechanisms for long-horizon task organization~\cite{chen2025robohorizon}. Fast-WAM studies whether future generation is needed at test time~\cite{yuan2026fast}, while AHA-WAM uses asynchronous world and action branches with observation-guided context routing~\cite{cai2026aha}. \method addresses a complementary system-level problem: how to update a high-level plan when an execution error or external disturbance invalidates a condition needed later in the task.

\subsection{Plan Verification and Recovery}

\nocite{ahmad2025unified}

Structured representations can support both plan checking and online recovery. VeriGraph uses scene graphs to verify and refine candidate action sequences~\cite{ekpo2024verigraph}. Other graph-based systems maintain scene information or task structure for planning and execution~\cite{shirasaka2025spark,yang2026integrated,zhang2026atomic}. LERa uses visual feedback to identify failures and revise a plan~\cite{pchelintsev2025lera}, Reflective Planning reassesses actions using predicted observations~\cite{feng2025reflective}, and behavior-tree approaches provide reactive execution and recovery mechanisms~\cite{ahmad2025unified,colledanchise2016behavior,iovino2022survey}. Recent language-based systems use visual or multimodal feedback to detect execution failures, explain plan--execution mismatches, or trigger corrective replanning~\cite{guo2024doremi,zhou2025code,liu2023reflect,duan2024aha,skreta2024replan,huang2022inner}. Symbolic planning provides explicit preconditions and effects that can be checked before execution~\cite{yang2025guiding,kwon2025fast,chen2025code,chu2025llm+,liu2023llm+}. Causal-link monitoring and plan repair provide a basis for reasoning about support relations from executed actions to pending actions. \method builds on these ideas rather than treating dependency tracking itself as a new planning principle. It specifically addresses support conditions established by completed subtasks that later become invalid, using visual state assessment to update explicit producer--consumer dependencies before selecting a local repair scope. Retained actions constrain the continuation, and the combined plan is validated before execution resumes. The experiments compare this framework with adapted verification and recovery baselines under a shared evaluation protocol.

\section{Method}
\method separates high-level task management from low-level control: it constructs, checks, and repairs plans, while the execution policy carries out individual subtasks. We use bridge assembly as a running example: the robot places supports, grasps a beam, and lays it across them.

\subsection{Problem Formulation}
At round $t$, $O_t$ is the visual observation and $S_t$ is the symbolic state estimated from visual interpretation, robot status, and actuator feedback. Given $S_0$ and goal $g$, the Operator Planner generates a reference plan $P=(o_1,\ldots,o_n)$ of grounded PDDL operators with specified preconditions and effects~\cite{aeronautiques1998pddl}. VAL checks the plan~\cite{howey2004val}, which is then compiled into a directed acyclic graph $G=(V,E)$.
\begin{align}
P&=\operatorname{PreparePlan}(S_0,g)=(o_1,\ldots,o_n), \nonumber\\
G&=(V,E)=\operatorname{BuildGraph}(S_0,P,g).
\label{eq:plan}
\end{align}
The nodes in $V$ are subtasks, \textsc{start}, and \textsc{goal}. Causal edges record which action supplies another's required condition; resource edges encode ordering constraints from resource use or modeled interference. Here, causal means PDDL condition support, not a learned model. The reference plan orders the graph and breaks scheduling ties.\looseness=-1

The selected operator is passed to the low-level policy as a natural-language instruction with the current visual observation. After execution, \method updates the state and checks the outcome and dependencies. A confirmed execution failure produces an error record $e_t$ and triggers plan repair:
\begin{equation}
(P',G')=\operatorname{TopologyGuidedRepair}(S_{t+1},e_t,P,G).
\end{equation}
The updated plan combines a repair fragment with retained unfinished operators. Validation from $S_{t+1}$ establishes consistency with the symbolic model and supplied state, without guaranteeing perception accuracy or physical success. Completed actions' valid effects remain in the current state; the completed prefix is not replayed.

\subsection{Pre-Execution Planning}
\begin{wrapfigure}{R}{0.48\textwidth}
  \centering
  \includegraphics[width=\linewidth]{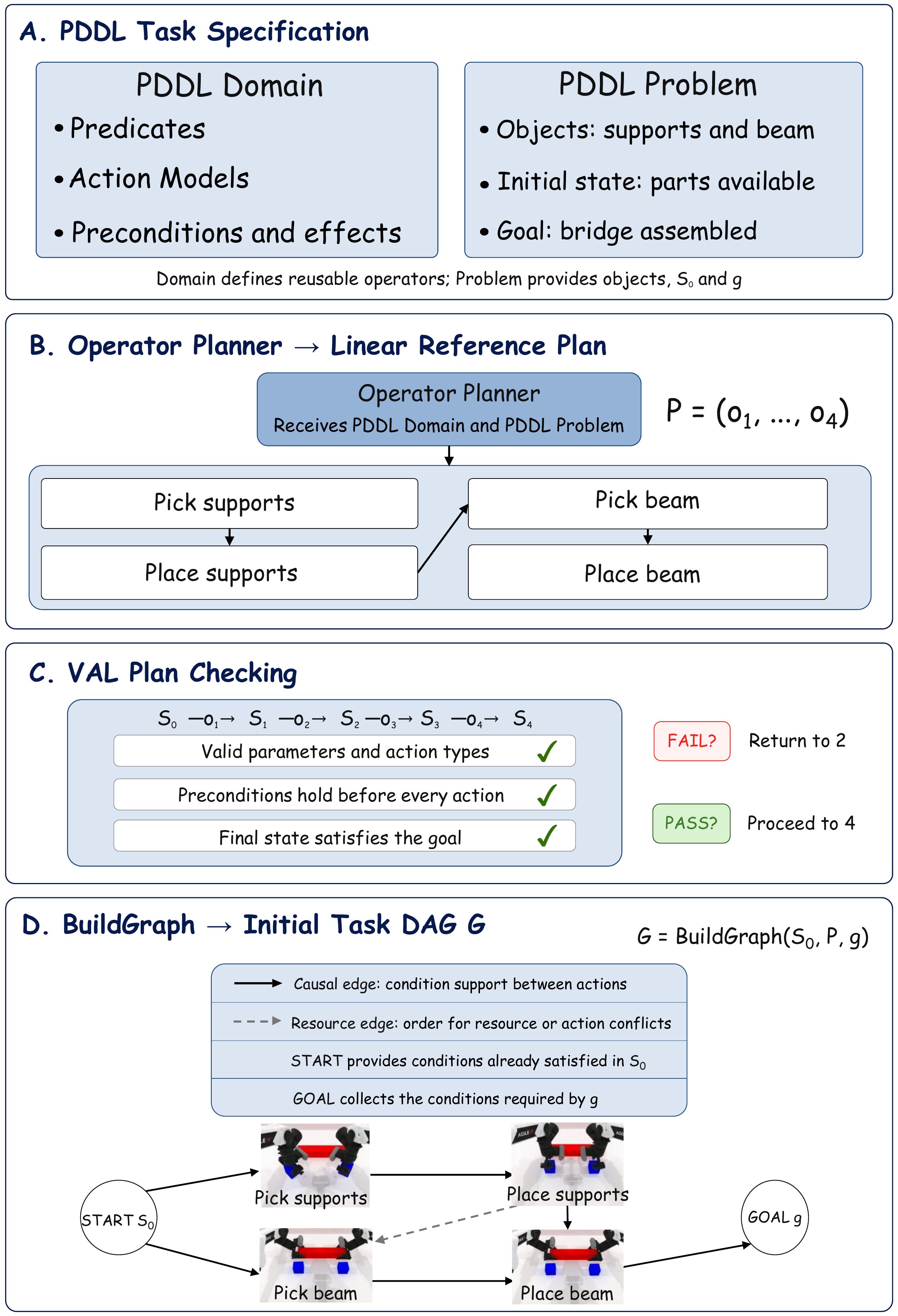}
  \caption{\method pre-execution planning and task-graph construction. The Operator Planner generates the reference plan $P$, where each $o_i$ in panel (B) is a grounded high-level PDDL operator. After validation by VAL, $P$ is compiled into the initial task graph $G$.}
  \label{fig:preplan}
\end{wrapfigure}
\textbf{Plan generation and validation.}
The Operator Planner receives a PDDL Domain and Problem (Fig.~\ref{fig:preplan}). The Domain specifies reusable operators, parameters, preconditions, and effects; the Problem specifies objects, initial state $S_0$, and goal $g$~\cite{aeronautiques1998pddl}. LAMA-first~\cite{richter2010lama} in Fast Downward~\cite{helmert2006fast} searches for a grounded sequence reaching $g$ from $S_0$. VAL independently checks action bindings, preconditions, modeled effects, and the final goal~\cite{howey2004val}. On failure, it returns the location and reason. The Domain, objects, $S_0$, and $g$ remain unchanged; the failed sequence is excluded before searching for an alternative grounding or ordering. Graph construction begins only after validation passes.

\textbf{Task graph construction.}
Each operator becomes a subtask node. For each precondition of $o_i$, the compiler searches preceding operators in reverse order for the nearest producer without an intervening invalidation in $P$. A causal edge links that producer to $o_i$ and is labeled with the supported condition. Conditions supplied by $S_0$ originate at \texttt{START}; goal conditions link to \texttt{GOAL} by the same rule. Placing a bridge support establishes the support-position condition needed by beam placement.\looseness=-1

Resource edges retain ordering for exclusive resources, such as an arm or gripper, and modeled interference. Our PDDL uses typed object parameters with deterministic logical preconditions and effects. Operators interfere when one invalidates a condition required or established by the other. Both edge types follow $P$, making the graph acyclic and $P$ a topological order. Other topological orders need not be valid; execution order is checked online.\looseness=-1

\begin{figure}[!htbp]
  \centering
    \includegraphics[width=0.98\linewidth]{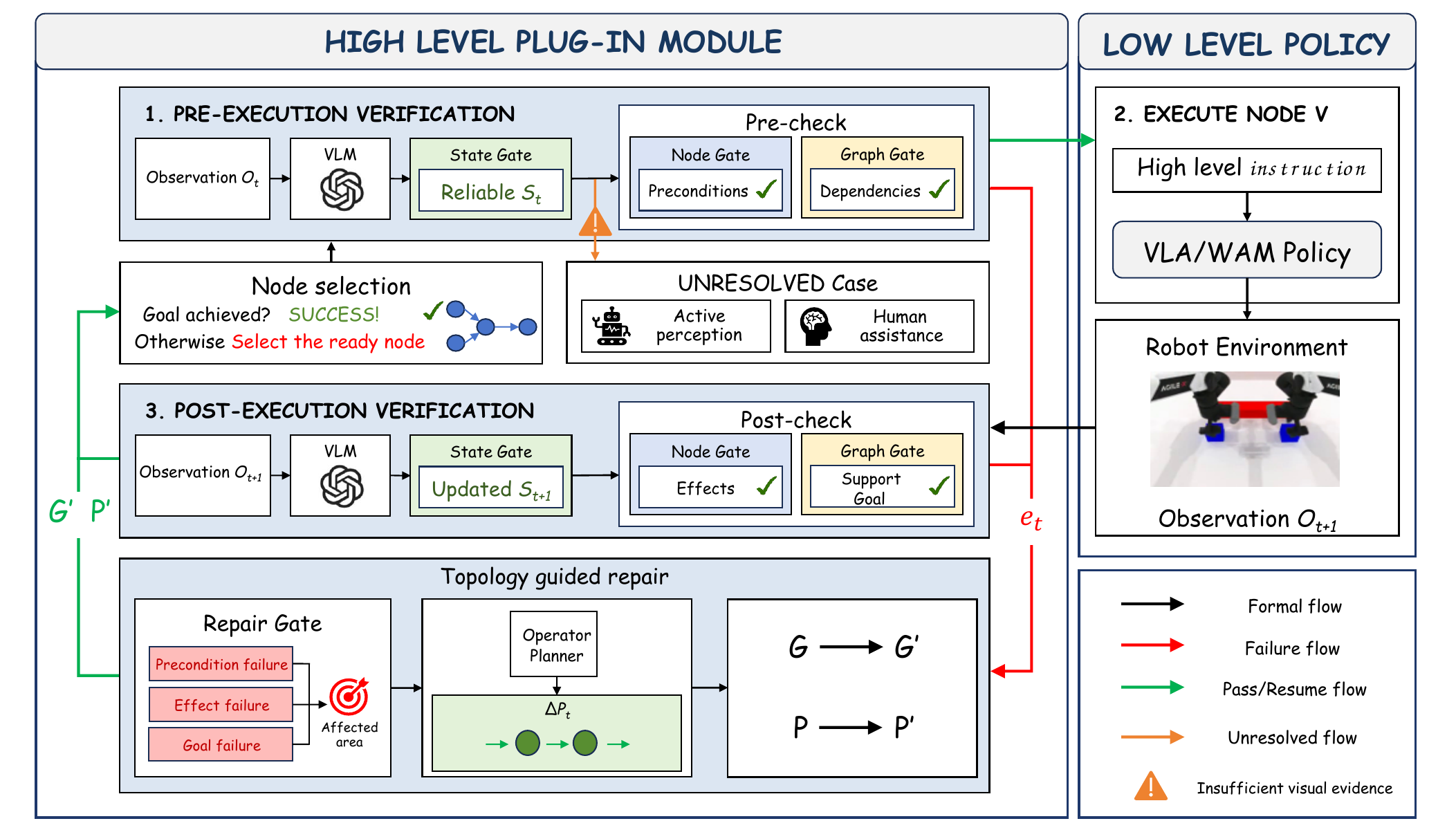}
    \caption{\method online verification and topology-guided causal local repair. The three gates establish reliable task facts, verify the selected subtask, and check graph dependencies. Upon a failure, the Repair Gate identifies affected nodes, validates a local repair plan, rebuilds the task graph, and resumes execution. Human assistance is reserved for future evaluation.}
  \label{fig:overview}
\end{figure}

\subsection{Online Verification and Recovery}
\textbf{Task-graph scheduling and online checking.}
As shown in Fig.~\ref{fig:overview}, the system schedules actions using causal dependencies and resource order in $G$. Among unfinished actions whose predecessors are complete, it selects the earliest action in $P$ whose execution conditions hold. If none is ready, it selects the earliest action with completed predecessors, refreshes the state, and checks whether repair is needed. Once all actions are complete, it checks $g$.

Before execution, the VLM extracts facts from $O_t$ about the current action and its dependencies. The State Gate combines them with robot state and actuator feedback to form $S_t$. If evidence is insufficient, the system acquires new images, adjusts the viewpoint, or repeats checks. Facts still unconfirmed within the observation budget yield \texttt{UNRESOLVED} and stop execution. The Node Gate checks preconditions; the Graph Gate checks resource predecessors and condition support. Confirmed unmet preconditions yield \texttt{FAIL} with \texttt{PRECONDITION\_FAILURE}. Otherwise, the operator is converted into an instruction for the low-level policy.

After execution, the system updates $S_{t+1}$ from $O_{t+1}$ and robot feedback. The Node Gate checks the expected effects and marks the node complete if they pass. The Graph Gate then checks whether support from \texttt{START} and completed actions still holds for remaining actions and \texttt{GOAL}. Missing effects or invalidated support yield \texttt{FAIL} with \texttt{EFFECT\_FAILURE}. Thus, a support displaced while grasping the beam triggers repair even if the grasp succeeds; the completed support-placement node is not reopened. If both checks pass, the system returns \texttt{PASS}. At \texttt{GOAL}, a verified $g$ yields success; a confirmed unmet goal yields \texttt{FAIL} with \texttt{GOAL\_FAILURE}. Algorithm~\ref{alg:logiv} in Appendix~\ref{app:algorithms} gives the execution loop, with these checks summarized as
\begin{equation}
\begin{aligned}
(\texttt{status},S_{t+1},e_t)
&=\operatorname{VerifyAndExecute}(v,G,S_t),\\
\texttt{status}
&\in\{\texttt{PASS},\texttt{FAIL},\texttt{UNRESOLVED}\}
\end{aligned}
\label{eq:verify_execute}
\end{equation}
The checked node is $v$. Record $e_t$ gives the failure type, node, unmet condition, and failed edge; it is empty otherwise. Completion status distinguishes missing action effects from a completed action's invalidated support.\looseness=-1

\textbf{Repair scope selection.}
Algorithm~\ref{alg:repair} in Appendix~\ref{app:algorithms} distinguishes affected actions from those that must be replaced. After a failure, we identify the unfinished action at which the failure was detected, or the unfinished actions that directly depend on invalidated support, together with their unfinished causal successors. Initially, only a just-executed action whose own effects failed is replaced. An action with a missing precondition is retained, and repair restores that condition before execution. If support from a completed action is later invalidated, the action remains complete and its unfinished consumers are retained. A final-goal failure starts with an empty replacement set. In the beam example, beam placement remains in the plan while repair restores its required support.

\textbf{Repair planning and validation.}
Once the replacement scope is fixed, the remaining unfinished actions form the retained plan $P_{\mathrm{keep}}$. The system works backward from the final goal $g$ through these actions to determine what must hold before their execution resumes; this result is denoted by $g_{\mathrm{rep}}$. For actions with conjunctive preconditions and goals and unconditional add/delete effects, conditions are represented as signed literals. The system processes the retained actions in reverse order, starting from $g$. If an action deletes a condition that is still required, the requirement is infeasible; otherwise, the system removes the conditions established by that action and adds its preconditions. Complementary literals also make the requirement infeasible and cause planning to fail immediately. If no actions are retained, the repair fragment must directly achieve $g$. For example, a retained beam-grasp action requires \texttt{gripper\_empty} and establishes \texttt{holding(beam)} for placement. The repair therefore needs an empty gripper before the grasp, but need not establish \texttt{holding(beam)} in advance. Starting from $S_{t+1}$, the Operator Planner generates a repair fragment $\Delta P_t$ that achieves $g_{\mathrm{rep}}$, giving $P'=\Delta P_t\oplus P_{\mathrm{keep}}$. The system uses VAL to check whether the complete plan $P'$ reaches $g$. A failure inside the repair fragment triggers replanning within the same scope, while a repeated candidate terminates repair. A failure at a retained action expands the replacement scope to that action and its unfinished causal successors. A planning failure, timeout, or unlocated failure invokes the deterministic fallback in the Notes. After validation succeeds, \method rebuilds the graph from $P'$: repair actions become new nodes, completed actions remain complete, and their still-valid effects are represented by the current state and the new \texttt{START} node. Execution resumes from the earliest ready node in reference order. Repair terminates when no further expansion is possible or the budget is exhausted, without guarantees of minimality or completeness.\looseness=-2

\section{Experiments}
\subsection{Experiment Setup}
\textbf{Tasks and perturbations.}
We evaluate \method on ten tasks from RoboTwin~V2~\cite{chen2025robotwin} and ten from BiCoord~\cite{peng2026BiCoord}. Both benchmarks support dual-arm manipulation with multistage actions and cross-arm dependencies. Appendix~\ref{app:tasks} lists the selected tasks in Table~\ref{tab:tasks}. Each task has ten predefined initial states. The clean setting evaluates each state once without an injected disturbance. The perturbed setting reuses the same initial states and introduces one recoverable disturbance at a predefined event. The disturbance type and trigger are fixed within each task and vary across tasks. For example, in \texttt{move\_can\_pot}, the pot is shifted laterally by 10–15 cm after the can is grasped and lifted.

\begin{figure}[!t]
  \centering
  \includegraphics[width=\linewidth]{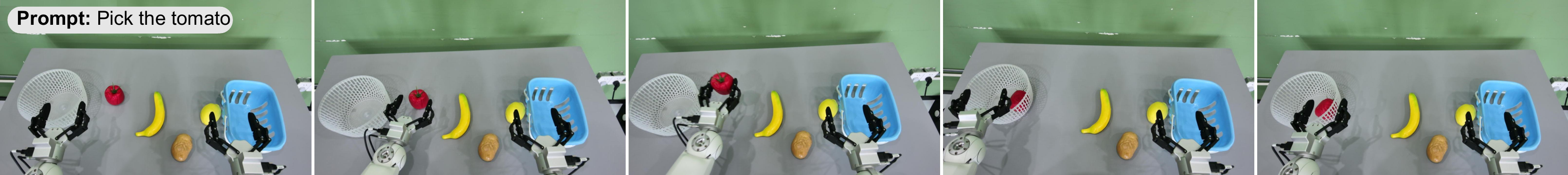}\\[6pt]
  \includegraphics[width=\linewidth]{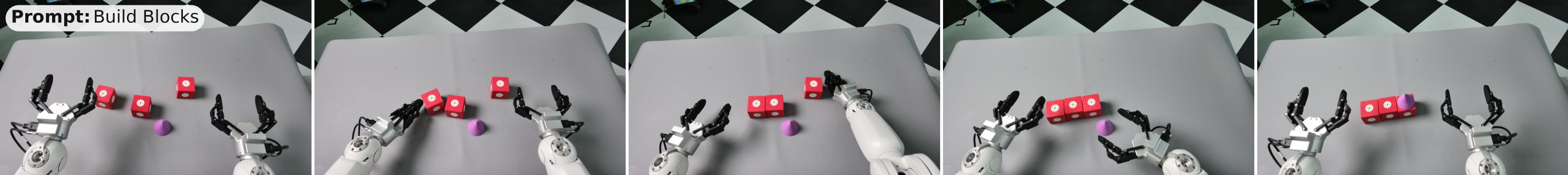}\\[6pt]
  \includegraphics[width=\linewidth]{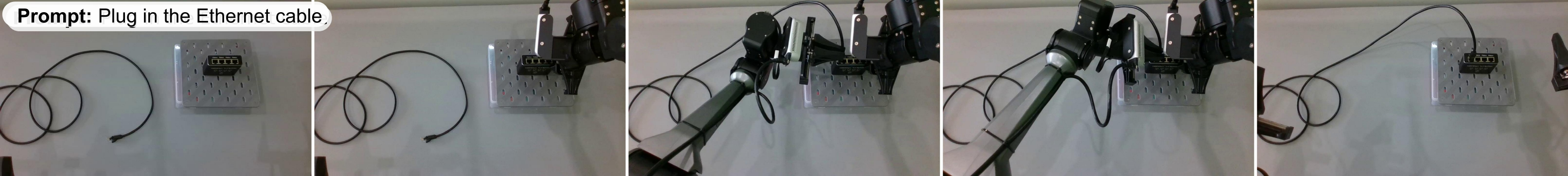}
  \caption{Real-robot rollouts of the three tasks. Top: pick the tomato and place it in the bowl. Middle: build the block structure with both arms. Bottom: plug in the Ethernet cable.}
  \label{fig:real}
\end{figure}

We design three real-world tasks (Fig.~\ref{fig:real}). In \texttt{pick\_tomato}, the robot moves a tomato from the table into a bowl, a short task with few conditions. In \texttt{build\_blocks}, both arms build a small block structure, where earlier placements establish support conditions for later ones. In \texttt{insert\_cable}, the robot plugs an RJ45 connector into a switch port, requiring millimeter-level alignment beyond pick-and-place. No disturbances are injected in these trials.\looseness=-1

\textbf{Backbones and baselines.}
We use $\pi_{0.5}$~\cite{intelligence2025pi_}, AHA-WAM~\cite{cai2026aha}, and Light-WAM~\cite{li2026light} as execution policies. For RoboTwin~V2 and BiCoord, each backbone undergoes task-specific fine-tuning on the ten selected tasks in that benchmark. The backbone training or adaptation uses expert demonstrations without the evaluation-time perturbations. LOGIV itself introduces no additional trainable components. For each backbone, we compare LOGIV with the backbone alone and three adapted recovery methods. VeriGraph-PDDL follows the scene-graph verification approach of VeriGraph~\cite{ekpo2024verigraph} and replans the remaining task from the current symbolic state when a constraint is violated. LERa-PDDL uses visual feedback to analyze a failure, updates the planning conditions, and replans the remaining task~\cite{pchelintsev2025lera}. VLM-RP-BT organizes the initial plan as a behavior tree~\cite{colledanchise2016behavior,iovino2022survey} and inserts recovery actions when a required condition is not satisfied. These names refer to the adaptations evaluated here, rather than to unmodified implementations of the original systems.

\textbf{Implementation details.} The experimental PDDL specification is predefined and held fixed. The VLM configuration uses GPT-4o with a maximum of 512 output tokens per call. Each node receives one pre-execution and one post-execution check. Human assistance is disabled in all reported experiments; potential failures first trigger active perception by the robot. An unresolved assessment allows at most three additional observations. We use LAMA-first in Fast Downward with a 30~s limit per planning call and a 5~s limit per VAL call. Each failure allows at most three local repair rounds and four VLM calls. Exceeding either limit terminates the episode. 

For these tasks, we compare the base $\pi_{0.5}$ policy with $\pi_{0.5}$ + \method on two physical dual-arm platforms: an Astribot~S1 whole-body dual-arm robot and a workcell built from two AgileX Piper 6-DoF arms with parallel-jaw grippers. Each platform provides a third-person RGB-D camera and a wrist camera per arm. The baseline and LOGIV configurations use the same $\pi_{0.5}$ checkpoint and the same demonstrations. \method adds no trainable components. Each configuration is evaluated over 10 trials per task under a shared protocol with matched initial layouts. A trial counts as successful only when the full instruction is completed.

\textbf{Evaluation metrics.}
We report full-task success without injected perturbations (C), full-task success with injected perturbations (P), and condition restoration (R). An event is eligible if execution reaches the injection point, the disturbance invalidates the designated condition, and recovery remains possible with the available action set. Restoration succeeds if the condition is restored within budget and passes the next relevant check. For each method, $R$ is the fraction of eligible events with successful restoration, counting failures in the denominator. Event sets may differ across methods, so $R$ is read alongside $P$. The ablation study additionally reports the fraction of unexecuted nodes deleted, replaced, or replanned during repair, and the number of still-valid completed actions executed again. These measures describe plan changes and redundant execution, respectively.

\subsection{Detailed Results}
We report results from simulation benchmarks and real-world robot experiments.

\begin{table}[!htbp]
\centering
\caption{\normalfont Task completion and condition restoration across backbones and online correction methods.}
\label{tab:main}
\small
\setlength{\tabcolsep}{5pt}
\renewcommand{\arraystretch}{1.08}

\begin{tabularx}{\linewidth}{@{}ll*{6}{>{\centering\arraybackslash}X}@{}}
\toprule
\multirow{2}{*}{Backbone} & \multirow{2}{*}{Method} & \multicolumn{3}{c}{\textbf{RoboTwin V2}} & \multicolumn{3}{c}{\textbf{BiCoord}} \\
\cmidrule(lr){3-5}\cmidrule(lr){6-8}
 & & C $\uparrow$ & P $\uparrow$ & R $\uparrow$ & C $\uparrow$ & P $\uparrow$ & R $\uparrow$ \\
\midrule
\multirow{5}{*}{$\pi_{0.5}$} & Backbone & 56 & 32 & 18 & 39 & 31 & 21 \\
 & VeriGraph-PDDL & 68 & 56 & 66 & 46 & 39 & 56 \\
 & LERa-PDDL & 67 & 57 & 64 & 47 & 43 & 63 \\
 & VLM-RP-BT & 63 & 58 & 68 & 45 & 45 & 59 \\
\rowcolor{gray!15}
\cellcolor{white} & \method & \textbf{72} & \textbf{60} & \textbf{72} & \textbf{50} & \textbf{48} & \textbf{68} \\
\addlinespace[5pt]
\multirow{5}{*}{AHA-WAM} & Backbone & 85 & 60 & 35 & 45 & 43 & 37 \\
 & VeriGraph-PDDL & 90 & 78 & 81 & 56 & 46 & 77 \\
 & LERa-PDDL & 88 & 79 & 82 & 52 & 48 & 73 \\
 & VLM-RP-BT & 91 & 80 & 80 & 57 & 48 & 80 \\
\rowcolor{gray!15}
\cellcolor{white} & \method & \textbf{93} & \textbf{83} & \textbf{85} & \textbf{62} & \textbf{53} & \textbf{84} \\
\addlinespace[5pt]
\multirow{5}{*}{Light-WAM} & Backbone & 74 & 58 & 28 & 43 & 39 & 31 \\
 & VeriGraph-PDDL & 76 & 68 & 65 & 56 & 46 & 67 \\
 & LERa-PDDL & 76 & 67 & 70 & 51 & 45 & 65 \\
 & VLM-RP-BT & 78 & 70 & 62 & 58 & 43 & 68 \\
\rowcolor{gray!15}
\cellcolor{white} & \method & \textbf{82} & \textbf{72} & \textbf{79} & \textbf{61} & \textbf{49} & \textbf{74} \\
\bottomrule
\end{tabularx}

\par\vspace{3pt}
\begin{minipage}{\linewidth}
\normalfont\small
\raggedright
\textbf{C}: Task success rate without artificial perturbations;
\textbf{P}: Task success rate after applying predefined perturbations;
\textbf{R}: Recovery success rate for invalidated conditions among valid, recoverable perturbation events.
All results are expressed as percentages (\%).
\end{minipage}
\end{table}

\textbf{Simulation experiments.}
Table~\ref{tab:main} summarizes the results of \method on both benchmarks. Without artificial perturbations, integrating LOGIV raises mean task success across the three backbones from 71.67\% to 82.33\% on RoboTwin~V2 and from 42.33\% to 57.67\% on BiCoord. Robots can still encounter state deviations and action failures in these runs. The gains therefore show that LOGIV improves normal task execution as well as recovery from injected disturbances. Under predefined execution-time perturbations, LOGIV also improves task completion. On RoboTwin~V2, task success with $\pi_{0.5}$, AHA-WAM, and Light-WAM reaches 60\%, 83\%, and 72\%, respectively, exceeding the corresponding backbone results by 28, 23, and 14 percentage points. On BiCoord, the respective success rates are 48\%, 53\%, and 49\%, with gains of 17, 10, and 10 percentage points. LOGIV achieves higher task success than all three correction baselines in both settings, indicating that its benefits extend across backbones and benchmarks. The perturbed results further show that the framework can help execution continue after an external disturbance. We also evaluate condition restoration separately from full-task completion. LOGIV achieves higher recovery success than the three correction baselines for every combination of benchmark and backbone, showing consistent gains in restoring invalidated conditions within the evaluation budget. These results support the use of dependency checks and plan repair to help execution continue after a failure. To guide repair, \method traces the conditions needed by subsequent actions, extending its assessment beyond the failed action itself. Retained actions constrain the updated plan, limiting changes to still-usable parts of the unfinished task.

\begin{table}[!htbp]
\centering
\caption{\normalfont Real-robot success over ten trials per task.}
\label{tab:real}
\normalsize
\renewcommand{\arraystretch}{1.12}
\setlength{\tabcolsep}{2.2pt}
\begin{tabular*}{\linewidth}{@{\extracolsep{\fill}}llcc@{}}
\toprule
Task & Platform & $\pi_{0.5}$ & $\pi_{0.5}$ + \method \\
\midrule
\texttt{pick\_tomato}  & Astribot S1  & 7 / 10 & \textbf{9 / 10} \\
\texttt{build\_blocks} & Astribot S1  & 3 / 10 & \textbf{5 / 10} \\
\texttt{insert\_cable} & AgileX Piper & 2 / 10 & \textbf{3 / 10} \\
\midrule
Overall & --- & 12 / 30 (40.0\%) & \textbf{17 / 30 (56.7\%)} \\
\bottomrule
\end{tabular*}
\end{table}

\textbf{Real-world experiments.}
Table~\ref{tab:real} reports the real-robot results. \method succeeds in 17 of the 30 trials compared with 12 for the base policy, raising the overall success rate from 40.0\% to 56.7\%. It improves performance on all three tasks. On \texttt{pick\_tomato}, successes increase from 7 to 9, with recovery from empty grasps or early releases of the tomato. On \texttt{build\_blocks}, successes increase from 3 to 5. This gain comes from restoring condition support: when a later reach displaces a previously placed block, the Graph Gate detects that a condition required by the remaining placements no longer holds. Repair restores that block while retaining placements that remain valid. On \texttt{insert\_cable}, successes increase only from 2 to 3. Most residual failures occur when the plug is aligned but fails to seat, so a successful retry depends on the contact-rich precision of the low-level policy rather than on the plan. With ten trials per task, wide confidence intervals indicate a trend rather than a precise effect size.\looseness=-1

\subsection{Ablation Study}
\begin{figure}[!t]
  \centering
  \includegraphics[width=0.70\linewidth]{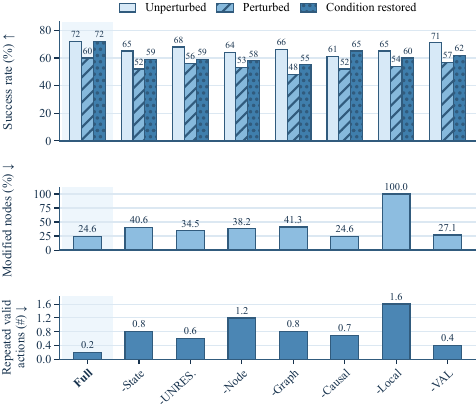}
  \caption{Ablation results of LOGIV. The top panel shows task completion and condition restoration under perturbations. The middle and bottom panels show modified nodes and repeated valid actions during repair, respectively.
  Full: complete method; -State: no State Gate; -UNRES: no uncertainty handling; -Node: no Node Gate; -Graph: no Graph Gate; -Causal: no causal dependency tracing; -Local: replanning all remaining actions instead of local repair; -VAL: no logical validation of the repaired plan.}
  \label{fig:ablation}
\end{figure}

We evaluate each ablation variant over 100 episodes per setting on the ten RoboTwin~V2 tasks in Appendix~\ref{app:tasks} (Table~\ref{tab:tasks}), using $\pi_{0.5}$ with shared visual inputs, VLMs, and planning budgets.

\textbf{Execution verification.}
Removing the State Gate, Node Gate, or Graph Gate lowers all three success measures. Removing the Graph Gate gives the lowest perturbed-task success, 48\% compared with 60\% for the full framework. This result is consistent with a benefit from checking dependencies beyond the selected action. Removing uncertainty handling lowers condition restoration from 72\% to 59\%. \looseness=-1

\textbf{Dependency tracing and local repair.}
The variant without causal tracing selects repair nodes by their positions in the linear plan. Its mean modified-node fraction remains 24.6\%, but clean-task success drops from 72\% to 61\% and perturbed-task success from 60\% to 52\%. Thus, the size of the selected region alone does not explain the result; the choice of nodes also matters. Replanning all remaining actions increases the modified-node fraction to 100\% and repeated valid actions from 0.2 to 1.6, while perturbed-task success decreases to 54\%. The modified-node fraction partly reflects this variant's definition and should be read alongside task success and repeated execution.\looseness=-1

\textbf{Plan validation.}
Without validation of the repaired plan, condition restoration decreases from 72\% to 62\%, while perturbed-task success decreases from 60\% to 57\%. VAL checks whether the repair fragment is compatible with retained operators and whether the complete continuation reaches the goal under the symbolic model. These results are consistent with a benefit from checking the combined plan rather than accepting a local fragment in isolation.\looseness=-1

\section{Conclusion}
We presented \method, a framework that connects symbolic task dependencies with observation-based execution checks and local plan repair. After detecting an invalidated condition, \method constructs a repaired continuation while retaining usable unfinished actions and validates the result before execution resumes. Experiments on RoboTwin~V2 and BiCoord show higher observed task success across three execution backbones, while the number of successful real-robot trials increases from 12/30 to 17/30. Ablations are consistent with benefits from dependency tracing, local repair, and validation. The findings are limited to the evaluated tasks, perturbations, and trial counts. Future work will examine broader disturbances, incomplete state information, and the runtime and resource costs of online recovery. We will also evaluate human assistance in failure recovery.\looseness=-1

\FloatBarrier
\bibliographystyle{abbrvnat}
\bibliography{references}

\clearpage
\useRomanappendicesfalse
\appendices
\section{Execution and Repair Algorithms}
\label{app:algorithms}
This appendix provides the pseudocode for the graph execution interface described in Section~III. The two procedures separate execution monitoring from plan repair: the first determines when the current graph can advance, and the second constructs a validated continuation after a confirmed failure.

\subsection{Task-Graph Execution}
Algorithm~\ref{alg:logiv} connects initial planning, graph-based scheduling, observation-based verification, and repair. It advances only when the selected action and its required support pass the checks; unresolved state estimates terminate the episode. Goal satisfaction is checked explicitly after all actions are complete.

\begin{algorithm}[!ht]
\algcaption{\method Task-Graph Execution and Repair}
\label{alg:logiv}
\textbf{Require} initial task state $S_0$, task goal $g$\par
\textbf{Output} \textsc{success} or \textsc{failure}\par\vspace{2pt}
\begin{compactalgorithmic}
\item $P\leftarrow\operatorname{PreparePlan}(S_0,g)$
\item \textbf{if} $P=\textsc{fail}$ \textbf{then return} \textsc{failure}
\item $G\leftarrow\operatorname{BuildGraph}(S_0,P,g)$; $t\leftarrow 0$
\item \textbf{while} execution budget remains \textbf{do}
\item \hspace*{1.5em}\textbf{if} all actions are complete \textbf{then} $v\leftarrow\texttt{GOAL}$
\item \hspace*{1.5em}\textbf{else} $v\leftarrow\operatorname{SelectNode}(G,P,S_t)$
\item \hspace*{1.5em}\textbf{if} $v=\varnothing$ \textbf{then return} \textsc{failure}
\item \hspace*{1.5em}$(\texttt{status},S_{t+1},e_t)\leftarrow\operatorname{VerifyAndExecute}(v,G,S_t)$
\item \hspace*{1.5em}\textbf{if} $\texttt{status}=\texttt{UNRESOLVED}$ \textbf{then return} \textsc{failure}
\item \hspace*{1.5em}\textbf{if} $v=\texttt{GOAL}$ and $\texttt{status}=\texttt{PASS}$ \textbf{then return} \textsc{success}
\item \hspace*{1.5em}\textbf{if} $\texttt{status}=\texttt{FAIL}$ \textbf{then}
\item \hspace*{3em}$(P',G')\leftarrow\operatorname{TopologyGuidedRepair}(S_{t+1},e_t,P,G)$
\item \hspace*{3em}\textbf{if} $P'=\textsc{fail}$ \textbf{then return} \textsc{failure}
\item \hspace*{3em}$(P,G)\leftarrow(P',G')$
\item \hspace*{1.5em}\textbf{end if}
\item \hspace*{1.5em}$t\leftarrow t+1$
\item \textbf{end while}
\item \textbf{return} \textsc{failure} \quad \algcomment{budget exhausted}
\end{compactalgorithmic}
\noindent\textit{Notes.} $\operatorname{SelectNode}$ uses Section~III-C's scheduling rule and returns $\varnothing$ if no unfinished action has completed predecessors; \texttt{START} counts as complete. $\operatorname{VerifyAndExecute}$ updates the state and $G$, marking the action complete once its effects pass, then checking remaining support.\looseness=-1
\end{algorithm}
\FloatBarrier

\subsection{Topology-Guided Repair}
Algorithm~\ref{alg:repair} distinguishes actions affected by a failure from actions that must be replaced. The retained unfinished actions determine the repair requirements through a backward pass. A candidate is accepted only after validation of the repair fragment together with the retained continuation; failures can expand the replacement scope within the stated budget. The procedure does not guarantee a minimal repair or completeness.

\begin{algorithm}[!ht]
\algcaption{\textbf{TopologyGuidedRepair}}
\label{alg:repair}
\textbf{Require} symbolic state $S_{t+1}$, error $e_t$, plan $P$, graph $G$\par
\textbf{Output} repaired plan and graph $(P',G')$, or \textsc{fail}\par\vspace{2pt}
\begin{compactalgorithmic}
\item $g\leftarrow\operatorname{Goal}(G)$; $(\mathcal{D},\mathcal{A})\leftarrow\operatorname{Locate}(e_t,G)$; $H\leftarrow\varnothing$
\item \textbf{while} repair budget remains \textbf{do}
\item \hspace*{1.5em}$P_{\mathrm{keep}}\leftarrow P_{\mathrm{unfinished}}\setminus\mathcal{A}$ \algcomment{retained plan}
\item \hspace*{1.5em}$g_{\mathrm{rep}}\leftarrow g$ \algcomment{repair goal}
\item \hspace*{1.5em}\textbf{for each} $a\in P_{\mathrm{keep}}^{\mathrm{rev}}$ \textbf{do}
\item \hspace*{3em}$g_{\mathrm{rep}}\leftarrow\operatorname{Backstep}(g_{\mathrm{rep}},a)$
\item \hspace*{3em}\textbf{if} $g_{\mathrm{rep}}=\textsc{fail}$ \textbf{then return} \textsc{fail}
\item \hspace*{1.5em}\textbf{end for}
\item \hspace*{1.5em}$\Delta P_t\leftarrow\operatorname{OperatorPlanner}(S_{t+1},g_{\mathrm{rep}})$; $u\leftarrow\varnothing$
\item \hspace*{1.5em}\textbf{if} $\Delta P_t\notin\{\textsc{fail},\textsc{timeout}\}$ \textbf{then}
\item \hspace*{3em}\textbf{if} $\Delta P_t\in H$ \textbf{then return} \textsc{fail}
\item \hspace*{3em}$H\leftarrow H\cup\{\Delta P_t\}$; $P'\leftarrow\Delta P_t\oplus P_{\mathrm{keep}}$
\item \hspace*{3em}$(\texttt{valid},f)\leftarrow\operatorname{VAL}(S_{t+1},P',g)$
\item \hspace*{3em}\textbf{if} $\texttt{valid}$ \textbf{then return} $(P',\operatorname{BuildGraph}(S_{t+1},P',g))$
\item \hspace*{3em}\textbf{if} $f\in\Delta P_t$ \textbf{then continue} \algcomment{retry same scope}
\item \hspace*{3em}$u\leftarrow\operatorname{EarliestFail}(f,P_{\mathrm{keep}})$
\item \hspace*{1.5em}\textbf{end if}; \textbf{if} $u=\varnothing$ \textbf{then} $u\leftarrow\operatorname{Fallback}(P_{\mathrm{keep}},\mathcal{D})$
\item \hspace*{1.5em}\textbf{if} $u=\varnothing$ \textbf{then return} \textsc{fail}
\item \hspace*{1.5em}$\mathcal{A}\leftarrow\mathcal{A}\cup\operatorname{Succ}^{*}_{G}(u)$; $H\leftarrow\varnothing$
\item \textbf{end while}; \textbf{return} \textsc{fail}
\end{compactalgorithmic}
\noindent\textit{Notes.} $\mathcal{D}$ and $\mathcal{A}$ denote affected and currently replaced actions. $\operatorname{Backstep}$ performs the backward update in Section~III-C and returns \textsc{fail} on conflict. $P_{\mathrm{unfinished}}$ follows the order in $P$; $P_{\mathrm{keep}}^{\mathrm{rev}}$ is $P_{\mathrm{keep}}$ in reverse. $\operatorname{EarliestFail}$ selects the earliest failing retained action. $\operatorname{Fallback}$ selects the first retained action in $\mathcal{D}$, otherwise the first in $P_{\mathrm{keep}}$, or $\varnothing$. $\operatorname{Succ}^{*}_{G}$ includes $u$ and its unfinished causal successors. $\operatorname{Locate}$ follows the scope-selection rules in Section~III-C; $\operatorname{BuildGraph}$ follows Section~III-B; $H$ stores current-scope candidates.
\end{algorithm}
\FloatBarrier

\section{Simulation Task Suite}
\label{app:tasks}
Table~\ref{tab:tasks} lists the ten selected tasks from each simulation benchmark. These are the same tasks used for the clean and perturbed settings in Section~IV-A; the ablations use the RoboTwin~V2 subset. Each task has ten predefined initial states, giving 100 episodes per benchmark and setting. Task names are retained as benchmark identifiers to make the evaluation scope explicit. The table lists the selected subset, not the complete benchmark collections.\looseness=-1

\setcounter{table}{0}
\renewcommand{\thetable}{A\arabic{table}}
\renewcommand{\theHtable}{appendix.\arabic{table}}
\begin{table}[!htbp]
\centering
\caption{The 20 dual-arm tasks from RoboTwin~V2 and BiCoord.}
\label{tab:tasks}
\small
\renewcommand{\arraystretch}{1.04}
\setlength{\tabcolsep}{2pt}
\begin{tabular*}{\linewidth}{@{\extracolsep{\fill}}ll@{}}
\toprule
\multicolumn{2}{c}{RoboTwin~V2~\cite{chen2025robotwin}} \\
\cmidrule(lr){1-2}
\texttt{beat\_block\_hammer} & \texttt{blocks\_ranking\_size} \\
\texttt{handover\_block} & \texttt{move\_can\_pot} \\
\texttt{open\_microwave} & \texttt{place\_dual\_shoes} \\
\texttt{stack\_blocks\_three} & \texttt{stack\_bowls\_three} \\
\texttt{stamp\_seal} & \texttt{turn\_switch} \\
\midrule
\multicolumn{2}{c}{BiCoord~\cite{peng2026BiCoord}} \\
\cmidrule(lr){1-2}
\texttt{cook} & \texttt{put\_objects\_cabinet} \\
\texttt{divide\_block\_tower} & \texttt{place\_plate\_and\_cup} \\
\texttt{build\_bridge} & \texttt{clean\_table} \\
\texttt{collect\_pens} & \texttt{match\_blocks\_with\_signs} \\
\texttt{jigsaw} & \texttt{exchange\_mics} \\
\bottomrule
\end{tabular*}
\end{table}
\FloatBarrier

\end{document}